\documentclass{article}

\usepackage{microtype}
\usepackage{graphicx}
\usepackage{caption}
\usepackage{subcaption}
\usepackage{xspace}
\usepackage{pifont}
\newcommand{\cmark}{\ding{51}}%
\newcommand{\xmark}{\ding{55}}%
\usepackage{booktabs} 

\usepackage{hyperref}

\newcommand{\pd}{\textsc{PopDescent}\xspace}

\newcommand{\Function}[2]{\FUNCTION{\textsc{#1}(#2)}}
\newcommand{\Call}[2]{\textsc{#1}(#2)}

\usepackage[accepted]{icml2024}

\usepackage{amsmath}
\usepackage{amssymb}
\usepackage{mathtools}
\usepackage{amsthm}

\usepackage[capitalize,noabbrev]{cleveref}

\theoremstyle{plain}

\theoremstyle{definition}

\theoremstyle{remark}

\usepackage[textsize=tiny]{todonotes}

\icmltitlerunning{Scrap your Schedules with PopDescent}

\begin{document}

\twocolumn[
\icmltitle{Scrap your schedules with PopDescent}



\icmlsetsymbol{equal}{*}

\begin{icmlauthorlist}
\icmlauthor{Abhinav Pomalapally}{equal,berk}
\icmlauthor{Bassel Mabsout}{equal,bu}
\icmlauthor{Renato Mancuso}{bu}
\end{icmlauthorlist}

\icmlaffiliation{berk}{University of California, Berkeley;}
\icmlaffiliation{bu}{Boston University CPS Lab}

\icmlcorrespondingauthor{Abhinav Pomalapally}{abhinav$\_$pomalapally@berkeley.edu}
\icmlcorrespondingauthor{Bassel Mabsout}{bmabsout@bu.edu}
\icmlcorrespondingauthor{Renato Mancuso}{rmancuso@bu.edu}

\icmlkeywords{Machine Learning, ICML}

\vskip 0.3in
]



\printAffiliationsAndNotice{\icmlEqualContribution} 

\begin{abstract}

In contemporary machine learning workloads, numerous hyper-parameter search algorithms are frequently utilized to efficiently discover high-performing hyper-parameter values, such as learning and regularization rates.
As a result, a range of parameter schedules have been designed to leverage the capability of adjusting hyper-parameters during training to enhance loss performance.
These schedules, however, introduce new hyper-parameters to be searched and do not account for the current loss values of the models being trained.

To address these issues, we propose Population Descent ($\pd$), a progress-aware hyper-parameter tuning technique that employs a memetic, population-based search.
By merging evolutionary and local search processes, $\pd$ proactively explores hyper-parameter options during training based on their performance.
Our trials on standard machine learning vision tasks show that $\pd$ converges faster than existing search methods, finding model parameters with test-loss values up to 18\% lower, even when considering the use of schedules.
Moreover, we highlight the robustness of $\pd$ to its initial training parameters, a crucial characteristic for hyper-parameter search techniques.

\end{abstract}

\section{Introduction}\label{introduction}


Today's machine learning methods almost entirely rely on gradient descent as a core optimization technique.
Many recent deep learning tasks, whether supervised or unsupervised, include Neural Network (NN) optimization with ever-growing parameter spaces.
These methods have multiple hyper-parameters (e.g., learning rate, regularization rate, batch size, model weights, etc.), all affecting training results significantly \cite{GDsaddlePoints, choromanska2015loss, saddles}.
Thus, the creation of hyper-parameter tuning frameworks is an extensive research topic, and numerous hyper-parameter tuning and meta-learning methods have been proposed to solve this problem~\cite{optuna, keras_tuner}.

\textbf{Existing hyper-parameter-tuning literature} falls into three main categories.
1) Given different hyper-parameter combinations, they fully train each model and evaluate the final results; this is computationally wasteful~\cite{EAGSRS}.
2) They select hyper-parameters based on their performance in the first few training epochs~\cite{keras_tuner}.
3) They adjust hyper-parameters based on a predetermined schedule~\cite{LRscheduleforfastergraddescent, li2017stochastic, LRtypesadvanced, LRscheduleBetter, usingLRscheduleForDeepLearning}.

\textbf{One important limitation shared across these methods is the lack of "active" hyper-parameter optimization, meaning they adjust hyper-parameters without taking into account the model's progress on solving the task at hand.}
Schedules, for example, are often curves described solely as a function of the number of gradient steps taken~\cite{inverseTime, exponential}.
These curves often introduce new hyper-parameters that increase the size of the search space (decay rate, decay steps, etc.).
We seek to replace these "static" schedules with a framework capable of taking into account model progress.

Evolutionary/genetic algorithms~\cite{EA, EAGSRS} utilize a population-based solution and "mutate" models by introducing noise to parameters, and then select the individuals with the highest fitness for each "generation."
Memetic algorithms combine evolutionary algorithms with a local search process (usually gradient descent) \cite{Moscato1999MemeticAA, hybridResponseFiltering, extraMutation, DANGELO2021136, Xue2021EvolutionaryGD}.
They bridge the gap between exploring more space based on random noise (evolutionary step) and efficiently exploiting efficient gradient computations (local step).
\textbf{Memetic algorithms are not currently utilized for hyper-parameter search}, but are rather used for specialized optimization problems.
Yet, they can be leveraged to solve the problem of actively combining black-box and gradient-based optimization.


We propose Population Descent ($\pd$), the first memetic algorithm designed for local hyper-parameter optimization.
The key idea in $\pd$ is to \emph{actively} choose how much to explore the parameter and hyper-parameter space v.s. exploit a specific location on the loss curve.
This decision is based on a normalized fitness score representing each model's progress after each iteration of local updates, and an $m$-elitist selection process to choose which models to keep and which to mutate.
We add a constantly adaptive learning and regularization rate for each iteration of gradient steps, helping accelerate or decelerate learning and generalization of each NN based on current progress.
Non-differential based mutations for these rates allow for simple randomization by adding Gaussian noise; we also add Gaussian noise to NN weights.

We show in our evaluations that $\pd$, uniquely designed for hyper-parameter search, yields better results than existing hyper-parameter tuning methods, even with the use of schedules.
$\pd$ also outperforms ESGD~\cite{ESGD}, the state-of-the-art memetic algorithm built for solving similar tasks.
To demonstrate \pd's effectiveness on real deep-learning workloads, we apply the algorithm to the FMNIST~\cite{fmnist}, CIFAR-10~\cite{cifar}, and CIFAR-100~\cite{cifar-100} classification benchmarks.
We also conduct ablation studies justifying the effectiveness of the choices made in $\pd$.
$\pd$ achieves better test and training loss on every experiment we performed while taking fewer total gradient steps. We claim the following contributions:

\raggedbottom
\begin{itemize}
  \item The first memetic algorithm designed for hyper-parameter search, $\pd$ is simple to implement and performs better than the more complex state-of-the-art memetic algorithm designed for machine learning tasks (ESGD).
  $\pd$ is also much more insensitive to changes to its own hyper-parameters, a key requirement for hyper-parameter search methods.
  \item $\pd$'s \emph{active} tuning allows us to scrap the use of parameter schedules through the use of an evolutionary step: learning/regularization rates, and NN weights are proportionally perturbed by Gaussian noise after \emph{each local search step for the duration of training}.
  \item A generic, open-source reference implementation of $\pd$ based on TensorFlow 2 which can be used directly in machine learning tasks.
\end{itemize}

\section{Population Descent}\label{method}

\renewcommand{\O}{\textbf{Optimized}}
\renewcommand{\o}{\text{optimized}}
\renewcommand{\P}{\textbf{Population}}
\newcommand{\f}{\textsc{Fitness}}
\renewcommand{\i}{\text{individual}}
\newcommand{\w}{\text{weak}}
\newcommand{\M}{\textbf{Mutated}}
\newcommand{\Wm}{\textbf{\textit{WeightedMultinomial}}}
\renewcommand{\r}{\text{replacement}}
\renewcommand{\S}{\textbf{Strong}}
\newcommand{\m}{\text{mutated}}
\renewcommand{\L}{\textbf{\textit{LogNormal}}}
\renewcommand{\l}{\textsc{Loss}}
\renewcommand{\b}{\text{batch}}
\newcommand{\T}{\textit{\textbf{Training}}}
\newcommand{\CV}{\textit{\textbf{CrossValidation}}}
\newcommand{\cv}{\textit{\textbf{CV}}}
\newcommand{\fcv}{\f_\cv}
\newcommand{\B}{\textbf{Batch}}

$\pd$ is a memetic algorithm, meaning it combines both meta-optimization and gradient-based optimization into a single scheme.
We define the pseudocode of $\pd$ in \textit{Algorithm~\ref{alg:pop_descent}} which we hereafter describe in detail.

\subsection{Algorithm Definition}\label{method:algorithm}

The goal of $\pd$ is to find an optimal set of parameters forming what we term an $\i$.
An individual is constructed from sets of parameters $\theta$, and hyper-parameters $\alpha$.
We search for individuals which maximize a user-provided $\f$ function.
These individuals are maximized on batches from a held-out \textbf{Test} distribution that remains unseen during the procedure. Namely $\i^* = \langle \theta^*, \alpha^* \rangle =$ $$\underset{\langle \theta, \alpha \rangle \in \textbf{Individuals}}{\sup} \mathbb{E}_{\b \sim \textbf{\textit{Test}}}\left[ \f(\langle \theta, \alpha \rangle, \b)\right]$$

However, since the $\textbf{\textit{Test}}$ distribution must remain unseen, we are forced to make use of available proxy data in the form of a $\T$ distribution and a $\CV$ distribution.
This is standard in machine learning workloads.
We do not make assumptions on the differentiability of the provided $\f$ function.
This allows one to use of common metrics of performance such as accuracy.
Since the dimensionality of the parameter space can be exceedingly large (such as with Neural Networks), we make use of a $\textsc{Local Update}$ function which can efficiently update $\theta$, containing the bulk of the parameters of every individual.
We assume that invocations of $\textsc{Local Update}$ maximizes the $\i$'s expected $\f$ over the $\T$ set.
An example of such a function is Stochastic Gradient Descent (SGD) as defined in Algorithm~\ref{alg:userdef}.
SGD makes use of gradient-backpropagation to update $\theta$ in linear time.
\textsc{Local~Update} minimizes a differentiable $\l$ as a proxy for maximizing $\f$ with respect to $\theta$.
However, the $\textsc{Local Update}$ function does not modify the $\alpha$ hyper-parameters such as learning rates, the regularization magnitudes.

In order to find the best hyper-parameters, $\pd$ takes an $m$-elitist approach by holding onto a candidate set of individuals called a $\P$.
In each iteration, The $m$ fittest individuals from the $\P$ are kept untouched ($m$-elite), while the weakest ($|\P| - m$) individuals are always replaced.
We then pick replacements from the $\P$ but bias our choices towards fitter individuals.
These replacements then go through a $\textsc{Mutate}$ operation provided by the user. The mutation magnitude depends on the fitness of the individual.
That is, we mutate individuals more when they perform poorly.
In a sense, the normalized $\f$ value allows the algorithm to be aware of progress made during optimization, and explore more space when that is more beneficial.
Throughout the algorithm, $|\P|$ remains invariant.

\begin{algorithm}[H]
\caption{\pd}\label{alg:pop_descent}
\begin{algorithmic}[1]
    \REQUIRE $\i : \theta \times  \alpha$
    \REQUIRE $\f : \i \times \B  \to [0,1]$
    \REQUIRE $\textsc{Converged} : \i  \to \{0,1\}$
    \REQUIRE $\textsc{Local Update} : \i \times \B  \to \i$
    \REQUIRE $\textsc{Mutate} : \i \times [0,1] \to \i$
    \REQUIRE $\textbf{Training} : \textit{Distr}[\B]$
    \REQUIRE $\textbf{CrossValidation} : \textit{Distr}[\B]$
    \REQUIRE $\P : \{\i, \ldots\}$
    \REQUIRE $m : \mathbb{N}$
    \\[0.2cm]
    \Function{Population Descent}{$\P, m$}
    \WHILE{ $\neg~\textsc{Converged}(\P)$ }
        
        \STATE $\b_\T \sim \T$
        \STATE $\O \gets \hfill$
        $\text{\hspace{0.3cm}}\{\Call{Local Update}{\i, \b_\T}\hfill$
        $\text{\hspace{0.4cm}}\mid \i \in \P\}$
        \STATE $\b_\cv \sim \CV$
        \STATE $\fcv(x) = \f(x,\b_\cv)$
        \STATE $\Wm \gets Pr(X = x) = \hfill$
            $\text{\hspace{0.3cm}}\begin{cases}
                x \in \O & \frac{\fcv(x)}{\sum_{o \in \O}{\fcv(o)}}\\
                x \not\in \O & 0
            \end{cases}$
        \STATE $\M \gets \emptyset$
        \STATE $\S \gets \O$
        \FOR{$1 \ldots (|\P| - m$)}
            \STATE $\w \gets \underset{\fcv}{\textsc{Minimum}}(\S)$
            \STATE $\S \gets \S / \{\w\}$
            \STATE $\r \sim \Wm$
            \STATE $\M \gets \M \cup\{\textsc{Mutate}(\hfill$
            $\text{\hspace{0.3cm}}\r,1-\fcv(\r))\}$
        \ENDFOR
        \STATE $\P \gets \S \cup \M$
    \ENDWHILE
    \STATE \textbf{return} $\underset{\fcv}{\textsc{Maximum}}(\P)$
    \ENDFUNCTION
\end{algorithmic}
\end{algorithm}

$\pd$ terminates when the user-defined $\textsc{Converged}$ function outputs 1 (line 1).
Then, at each iteration:
\begin{enumerate}
    \item Lines 3-4: The individuals in the $\P$ all take a \textsc{Local Update} step over a batch sampled form the $\T$ distribution. This produces a set of $\O$ individuals;
    \item Lines 5-6: A batch is sampled from the $CV$ distribution, upon which we build $\fcv$, i.e., the fitness function for that batch;
    \item Line 7: We use $\fcv$ to build a $\Wm$ probability distribution, whose samples are individuals from the $\O$ set.
        The probability of each individual is defined by normalizing their fitness values, so that the probabilities sum to 1.
        This distribution is biased towards choosing fitter replacements;
    \item Line 8-15: We iterate ($|\P| - m$) times replacing the ($|\P| - m$) lowest fitness individuals by a mutated replacement.
        We find replacement individuals via sampling from the $\Wm$ distribution (Line 12).
        Then the replacement is mutated by an amount dependent on its fitness: the lower the fitness, the more it will be mutated;
    \item Line 16: $\P$ is now updated to include the $m$ $\S$ individuals and the ($|\P| -m$) mutated replacements;
    \item Line 18: Finally, we return the individual in the $\P$ with the largest fitness.
\end{enumerate}

\begin{algorithm}[H]
\caption{Example function implementations}\label{alg:userdef}
\begin{algorithmic}
    \REQUIRE $\l : \i \times \B  \to \mathbb{R}$
    \REQUIRE $\beta_1, \beta_2 : \mathbb{R}$
    \\[0.2cm]
    \Function {Local Update}{$\i, \B$}
        \STATE $\o \gets \i$
        \STATE $\o_\theta \gets \i_\theta\;+\hfill$
        $\text{\hspace{0.3cm}}\i_\alpha\nabla_{\i_\theta} \l(\i, \B)$
        \STATE \textbf{return} $\o$
    \ENDFUNCTION\\[0.2cm]
    \Function {Mutate}{$\i, \text{magnitude}$}
        \STATE $\m \gets \i$
        \STATE $\m_\theta \sim \textbf{\textit{Gaussian}}(\i_\theta, \beta_1\text{magnitude})$
        \STATE $\m_\alpha \sim \L(\i_\alpha, \beta_2\text{magnitude})$
        \STATE \textbf{return} $\m$
    \ENDFUNCTION
\end{algorithmic}
\end{algorithm}

In the example function implementations in Algorithm~\ref{alg:userdef}, we also show a sample $\textsc{Mutate}$ function where we randomize the $\theta$ parameters via a \textbf{\textit{Gaussian}} distribution whose standard deviation is defined by the mutation magnitude.
We opt to modify the learning rate geometrically via a \textbf{\textit{LogNormal}} distribution so that multiplying the learning rate by 0.1 and 10 is equally as likely with a standard deviation of 1.
Note that when the magnitude is at 0, none of the parameters would change.

\subsection{Key points in $\pd$'s design}
\textbf{Incentivizes Generalization:} We designed $\pd$ to naturally select individuals which generalize well to a dataset hidden during local updates.
We hypothesize that this would allow proper selection of regularization values rather than using ad-hoc techniques such as early stopping.
This is evaluated in \textit{Section~\ref{eval:ablation}}.

\textbf{Relates to Random Search:} If we remove the selection and mutation procedure, then $\pd$ simply becomes random search, since after random hyper-parameter initialization, the individuals only undergo iterations of SGD.

\textbf{Parallelizable:} Local updates to each individual in the population can occur in parallel with
the only synchronization required occurring in the replacements step.

\textbf{Insensitive:} $\pd$ has a few hyper-parameters itself (depending on the implementation of \textsc{Mutate}), but we have left these values constant across our experiments to showcase the effectiveness of the method and its low sensitivity to specific values of these parameters.

\subsection{Limitations}
$\pd$, as any optimization framework, is fundamentally constrained by the \emph{no free lunch theorem}~\cite{nofreelunch}, therefore there will always be a case where this algorithm will be worse than random at maximizing our $\f$.
For example, when the regularization rate is initialized too high for the whole population, a large number of randomization steps would be needed before making progress, due to the mutation method used being a random geometric walk.
We therefore leave parameter initialization and the mutation step redefinable with defaults that have been successful in all our machine learning tasks.
Another limitation is that the algorithm does not take into account the best individual ever observed, meaning there is no guarantee that the algorithm will always improve in performance with more iterations.
We make this trade-off actively choosing to always take a \textsc{Local Update} with respect to the whole $\P$ motivated by maximizing efficiency.
In our evaluations we did not observe significant performance degradation on the test-set occuring over iterations, which we discuss in \textit{Section~\ref{eval:convergence}}.

\section{Evaluations}\label{eval}

This section demonstrates that \textbf{1)}~$\pd$ achieves better performance than existing tuning/scheduling algorithms on the FMNIST, CIFAR-10, and CIFAR-100 vision benchmarks.
\textbf{2)}~While significantly simpler, $\pd$ converges faster than a problem-specific memetic algorithm in a fair comparison.
\textbf{3)}~$\pd$'s specific randomization scheme contributes to its results.
\textbf{4)}~$\pd$ is notably insensitive to changes in its own hyper-parameters, allowing it to tune target parameters without having to tune itself.

\newcommand{\barfigwidth}{0.3\textwidth}

\begin{figure*}[h]
     \centering
     \begin{subfigure}[b]{\barfigwidth}
         \centering
         \begin{subfigure}[b]{\textwidth}
             \includegraphics[width=\textwidth]{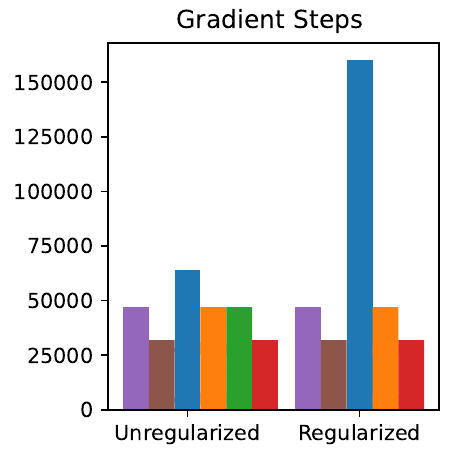}
             \includegraphics[width=\textwidth]{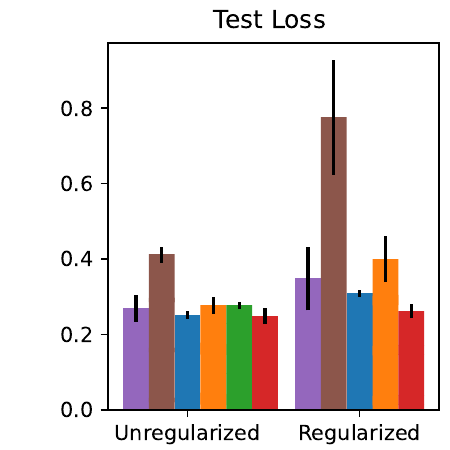}
         \end{subfigure}
         \caption{FMNIST}
         \label{fig:fmnist}
     \end{subfigure}
      \hfill
     \begin{subfigure}[b]{\barfigwidth}
         \centering
         \begin{subfigure}[b]{\textwidth}
             \includegraphics[width=\textwidth]{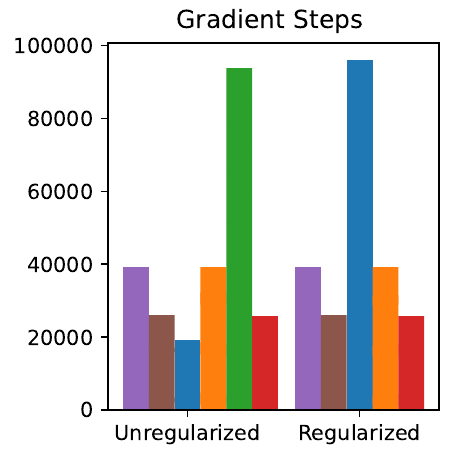}
             \includegraphics[width=\textwidth]{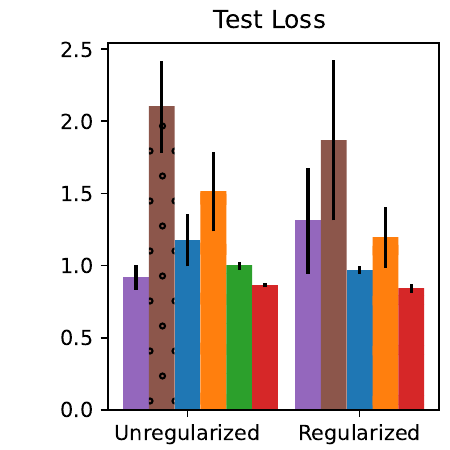}
         \end{subfigure}
         \caption{CIFAR-10}
         \label{fig:cifar10}
     \end{subfigure}
     \hfill
     \begin{subfigure}[b]{\barfigwidth}
         \centering
         \begin{subfigure}[b]{\textwidth}
             \includegraphics[width=\textwidth]{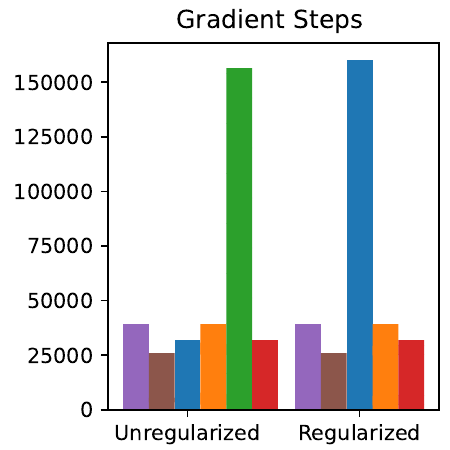}
             \includegraphics[width=\textwidth]{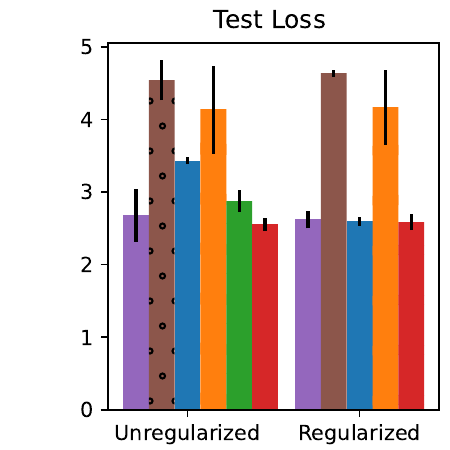}
         \end{subfigure}
         \caption{CIFAR-100}
         \label{fig:cifar100}
     \end{subfigure}
     \hfill
     \begin{subfigure}[b]{\textwidth}
        \centering
        \includegraphics[width=\textwidth]{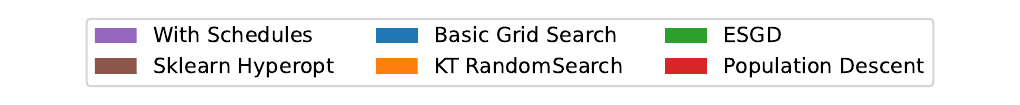}
     \end{subfigure}
     \vspace{-0.7cm}
     \caption{\textbf{Benchmark Tests.} The best test loss achieved by each method (the lower plots), plotted with standard deviation across 10 seeded trials;
     we show how many gradient steps each method takes to converge (the top plots).
     Each column of plots represents one vision benchmark, and we compare all methods' performance with and without a regularization term on each benchmark.
     $\pd$ (red bar with circles) achieves the lowest test loss in each problem.
     Table with quantitative data provided in the Appendix in \textit{Table~\ref{table:Bench}}.}
     \label{fig:Benchmarks}
\end{figure*}

\subsection{Benchmarks}\label{eval:bench}
We compare $\pd$ to a wide range of hyper-parameter tuning methods on standard machine learning vision benchmarks FMNIST, CIFAR-10, and CIFAR-100.
All frameworks search for the optimal learning and regularization rates, and they use a model's performance on an unseen/cross-validation dataset as their ranking heuristic.
A hidden test set for final evaluations is used as the objective.

In \textit{Figure \ref{fig:Benchmarks}}, we calculate the total number of gradients steps taken by every algorithm via $\textit{total} = \textit{iterations} \times \textit{model count} \times \textit{batches}$.
We run each method until convergence.

\textit{"Regularized"} tests add an L2 kernel regularizer to one layer of an identical CNN used in the \textit{"Unregularized"} tests.
In each problem, all frameworks search for the optimal learning and regularization rates.

\subsubsection{Competing Algorithms}\label{eval:competing}
\textbf{Grid Search}~\cite{EAGSRS} is the simplest and most commonly used search method.
We choose an array of possible values to choose from for each hyper-parameter.
Grid search then initializes a model for each possible combination of hyper-parameters, trains them fully, and evaluates the best hyper-parameters based on the final test results.
We test with five learning rates and five regularization rates, training 25 models to completion.

\textbf{Keras Tuner (KT) RandomSearch}~\cite{keras_tuner} and Sklearn's Hyperopt are two of the most popular available tuners (along with Optuna, which is almost identical).
We use KT's RandomSearch, a variation of a grid search.
It randomly samples hyper-parameter combinations (we evaluate 25 combinations) from a range of values instead of a discrete space like a regular grid search.

KT trains each combination for a few epochs (two in our case), evaluating the validation loss achieved before fully training each model.
Then, KT returns the hyper-parameters (not a trained model) that performed best at the end of the search.
By not training each model fully, RandomSearch efficiently samples more combinations than grid search.
We then train the model with the given hyper-parameters and employ early stopping.

\textbf{Sklearn Hyperopt TPE}~\cite{hyperopt} is identical to KT RandomSearch except in how it chooses which hyper-parameters to sample.
Instead of picking from a random distribution, Hyperopt uses a Tree-structured Parzen estimator (TPE), a form of Bayesian search.
Based on the last round's performance, TPE builds a probability model of the "best" hyper-parameter options to test next.
We also make use early stopping while running TPE.
We test TPE to evaluate $\pd$ against Bayesian methods.

\textbf{Schedules.} We compare against an exponential decay, inverse time decay, and polynomial decay learning rate schedule, very common choices for CNN optimization~\cite{inverseTime, exponential}.
These are based on the number of gradient steps taken, and not the model's performance.
Each schedule adds extra hyper-parameters to search (decay rate/decay steps).
Deciding which schedule to use also increases the search space.
To search these added parameter choices, we chose the tuning framework that performed best in our tests (\textit{Figure~\ref{fig:Benchmarks}}), which is KT RandomSearch.
KT randomly sampled any of the three schedules in addition to their respective initialization parameters, and yielded the best overall combination.
We agian train with early stopping.

\textbf{ESGD.} 
Evolutionary Stochastic Gradient Descent~\cite{ESGD}, is a state-of-the-art memetic algorithm.
It combines gradient descent with an evolutionary step, mutating the weights of individuals in a population of models.
ESGD does not specifically search for hyper-parameters, though it is a high-performing memetic algorithm, and thus a suitable competitor to $\pd$.
To compare fairly against the open-source implementation of ESGD which does not implement regularization, we exclude comparisons with regularization.

\subsubsection{Discussion}\label{discussion}

\textbf{$\pd$} (rightmost red bar with circles in each bar graph) achieves the lowest test loss in all scenarios while given the least amount of gradient steps (except in CIFAR-10 without regularization, where grid search converges quicker, but achieves 1.36x worse test loss than $\pd$), as seen in each of the bottom three figures in \textit{Figure~\ref{fig:Benchmarks}}.
A tabular representation of the results is provided in \textit{Section~\ref{table:Bench}} of the appendix.

\textbf{Grid search} (blue bar) when only searching for learning rates (Unregularized), is within $1.01\times$ $\pd$'s loss on FMNIST, but $1.36\times$ worse on CIFAR-10.
This inconsistency is explained by being the user's responsibility to estimate satisfactory hyper-parameter options for grid search to evaluate.
Also, grid search suffers the curse of dimensionality, with gradient steps taken exploding when searching for just 5 learning rates and 5 regularization rates (it took $5\times$ more gradient steps than $\pd$ to train CIFAR-10 with regularization).

\textbf{KT RandomSearch and Sklearn Hyperopt} (orange bar with stars, brown bar with dots) choose the "best" hyper-parameters based on a few early training epochs, struggling to make progress in later epochs.
Again, this leaves room for fine-tuning; either tuner's best test loss on any dataset plateaued at $1.1\times$ higher than $\pd$'s (KT RandomSearch on Unregularized FMNIST), while $\pd$ excels by customizing hyper-parameters at each epoch.

\textbf{Schedules} (purple bar with stripes) achieve test loss within 1.01x of $\pd$ on CIFAR-100 with regularization.
However, because they depend only on the number of gradient steps taken, they fail to consistently take advantage of drastic changes in loss values. For example, Scheduling performs $1.33\times$ worse than $\pd$ on FMNIST with regularization, and $1.55\times$ worse on CIFAR-10 with regularization.
Also, schedules add to search-space complexity with extra hyper-parameters to tune.

\textbf{ESGD} (green bar with stripes) required manual tuning of its learning rate and training batch size to properly converge, and still did not reach within $1.1\times$ $\pd$'s best model's test loss in any problem.
This highlights $\pd$'s advantage over problem-specific algorithms in addition to hyperparameter tuners, as $\pd$ automatically adjusts hyper-parameters during training, reaching convergence on its own.


\subsection{Convergence}\label{eval:convergence}
Memetic algorithms like ESGD often rely on mutation lengths, reproductive factors, mixing numbers, etc.
Essentially, their genetic schemes are complex: they introduce new hyper-parameters, and thus high implementation complexity.
This is undesirable for hyper-optimization.

\begin{figure}[t]
\begin{center}
\graphicspath{ {./Progress Graphs/} }
\centering
\includegraphics[width=0.5\textwidth]{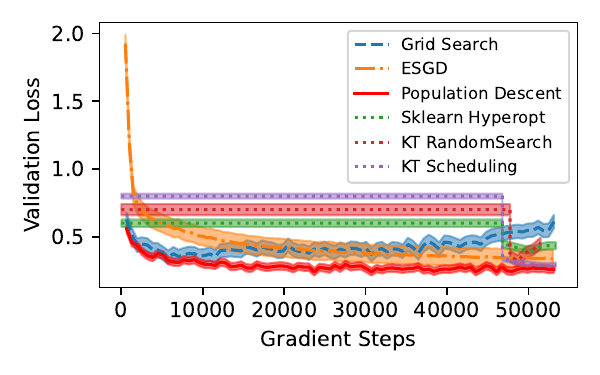}
\vspace{-1.2cm}
\end{center}
\caption{\textbf{Convergence.} A comparison of hyper-optimizers showcasing their validation loss progress against the number of gradient steps taken on the FMNIST dataset.
Each algorithm's exponential moving average across six random seeds is plotted with standard deviation.
They are tuning the learning rate (without regularization).
Note that Sklearn Hyperopt, KT RandomSearch, and KT Scheduling remain flat until 46k gradient steps.
Their tuning process evaluates each hyperparameter combination on two epochs at a time; real training occurs after this search (post 46k steps).
See \textit{Table~\ref{table:Convergence}} for quantitative results.
}
\label{figure:Fprog}
\end{figure}

\textbf{Simpler Design.} On the other hand, $\pd$'s mutation step only adds independent noise to parameters and uses a simple rank-based ($m$-elitist) recombination step.
Still, when comparing convergence of the highest fitness model in the population, as depicted in \textit{Figure~\ref{figure:Fprog}}, $\pd$'s final validation loss is $0.77\times$ that of the next best method's loss.

In \textit{Figure~\ref{figure:Fprog}}, we train each algorithm on six random seeds, running them for more iterations than optimal to show convergence/divergence over time (Grid Search for 100 iterations, KT RandomSearch for 25, KT Scheduling for 25, Sklearn Hyperopt for 32, \pd\ for 115, and ESGD for 15).
We plot the mean exponential moving average (bold line) of the cross-validation loss of the best model for each algorithm across all seeds, along with the standard deviation (shading), as a function of gradient steps taken.

Grid search (blue curve), Sklearn Hyperopt (green curve), and KT RandomSearch (dark red curve) all demonstrate overfitting.
Even though most use early stopping, their validation loss rises close to 50k gradient steps.
$\pd$ (bright red curve) on-average converges to 0.260 validation loss, and ESGD (orange curve) converges to 0.336, $1.29\times$ worse than $\pd$.
$\pd$ not only achieves a lower validation loss with lower standard deviation, but also requires fewer tunable hyper-parameters to implement its global step.
Additionally, $\pd$ converges quicker and its validation loss never evidences overfitting, in-part due to using cross-validation loss as our evolutionary ranking heuristic.

\subsection{Ablation Study}\label{eval:ablation}
We analyze how \textbf{1)} the \textit{randomization scheme} of NN weights/learning rate/regularization rate, and \textbf{2)} the use of \textit{cross-validation loss to evaluate the fitness} of individuals affect \pd's performance.
To emphasize differences, we add l2 kernel regularization to each layer in the benchmark FMNIST model, and reduce the training set size to 10K.
All tests are initialized with default learning/reg. rates of 0.001.
$|\P| = 10$, and $m = 5$.

First, we show how \pd's randomization scheme (NN weights, learning, and regularization rates) reduces the test loss to $0.84\times$ when turned on.
Adding Gaussian noise and choosing models that perform well on cross-validation loss helps the models explore more space, while selecting models that prevent overfitting.

Second, we evaluate how our use of normalized cross-validation loss as our fitness function acts as a heuristic for generalization.
Even on a model without regularization, cross-validation selection allows $\pd$ to achieve a test loss that is $0.31\times$ than that of selection done via a training set loss, illustrating how $\pd$'s selection scheme prevents even unregularized models from overfitting.
We present the most pronounced differences in \textit{Table~\ref{table:FA}} to best highlight \pd's features.


\subsection{Hyper-parameter Sensitivity}\label{eval:pd sensitivity}

In this section, we show that $\pd$ is less sensitive to its initialized local search parameters, relying less on user inputs and more on its evolutionary step to identify optimal parameters.
We show how varying local search parameters affects ESGD (a state-of-the-art memetic algorithm) more than \pd\ on the CIFAR-10 dataset.
We run each algorithm with a constant seed and constant hyper-parameters except one (either learning rate or the number of iterations).
One iteration defines one local and global update together.
A gradient update is taken each time before performing a mutation.

\begin{table}[t]
\centering

\vspace*{2mm}
\begin{tabular}{c|c|c}
\toprule
{\textit{Learning Rate}} & {\textit{Iterations}} & {$\textit{Test Loss} \pm \sigma$} 
\\
\midrule
\multicolumn{3}{c}{\textbf{Everything Constant Except Learning Rate}} \\
\midrule
\textbf{[0.01, 0.05, 0.001]}  & 30 & $1.049 \pm 0.172$\\ 
\midrule

\multicolumn{3}{c}{\textbf{Existing Constant Except Total Iterations}} \\
\midrule
0.001  & \textbf{[10, 30, 50]}  & $0.958 \pm 0.191$\\
\bottomrule
\end{tabular}
\caption{\textbf{$\pd$} training with variable local parameters. $\pd$ defaults to a batch size of 64, a learning rate of 0.001 with Adam, and 30 iterations for the CIFAR-10 benchmark.}
\label{table:PDLocalSens}
\end{table}

\begin{table}[t]

\centering

\vspace*{2mm}
\begin{tabular}{c|c|c}
\toprule
{\textit{Learning Rate}} & {\textit{Iterations}} & {$\textit{Test Loss} \pm \sigma$}\\
\midrule
\multicolumn{3}{c}{\textbf{Everything Constant Except Learning Rate}} \\
\midrule
\textbf{[0.01, 0.05, 0.001]}  & 3 & $1.325 \pm 0.582$\\ 
\midrule

\multicolumn{3}{c}{\textbf{Everything Constant Except Total Iterations}} \\
\midrule
0.001  & \textbf{[1, 3, 5]}  & $1.159 \pm 0.455$\\ 
\bottomrule
\end{tabular}
\caption{\textbf{ESGD} training with variable local parameters.
ESGD defaults to a batch size of 8, a learning rate of 0.01 with SGD, and 3 iterations for the CIFAR-10 benchmark (\pd\ trains over 128 batches per iteration, ESGD over the whole training set).}
\label{table:ESGDLocalSens}
\end{table}

\textit{Tables \ref{table:PDLocalSens}} and \textit{\ref{table:ESGDLocalSens}} show how when varying ESGD's learning rate, its standard deviation of results is $3.38\times$ higher than $\pd$'s, and when varying how many iterations we run each algorithm, ESGD's standard deviation is $2.38\times$ higher than $\pd$'s.
$\pd$ also has a much lower test loss across trials (avg. 19.2\% lower).
Complex memetic algorithms such as ESGD have a high number of adjustable hyper-parameters, and their performance depends significantly on their specific values.
As long as the parameters chosen are not extreme, these results show that the specificity of $\pd$'s initial hyper-parameters is not particularly effectful.

\textbf{Not Tuning the Tuner.} Another supporting note is how we never needed to change $\pd$'s hyper-parameters when evaluating different tasks, while still yielding the best model loss.
All tests for \pd\ across this entire paper (except the ablation tests) use the same population size (5) and randomization scheme (same Gaussian noise distributions) for the global step, and the same default learning rate (0.001), reg. rate (0.001), and batch size (64) for the local step (except during this experiment).


\section{Related works}\label{related}

\textbf{Gradient-Based Optimizers} such as Stochastic Gradient Descent~\cite{SGD} offer quick convergence on complex loss spaces.
As an improvement to SGD, momentum-based optimizers like Adam~\cite{AdamMomentum, adadelta} better traverse loss landscapes via utilizing momentum for escaping plateaus (which are prominent in high-dimensional loss spaces) and improving stability.
Adam's weight decay term also limits exploding gradients, and acts as a regularizer term.
Other options like the newer Sharpness-Awareness Minimization (SAM) or Shampoo optimizer, which use "preconditioning matrices," promise even faster convergence~\cite{Shampoo}.
\pd\ tunes hyper-parameters for \emph{any} local optimizer, including SGD, Adam, SAM, or Shampoo.
Such works are thus orthogonal to hyper-parameter optimization schemes.

\textbf{Hyper-parameter Search Methods:}
\textit{Grid search} is the most commonly used method for searching the hyper-parameter space due to its simplicity and generally understood performance for NN training.
Essentially, it is an exhaustive search of the Cartesian product of hyper-parameters.
It suffers the curse of dimentionality, i.e. it takes exponentially more gradient steps to train the more hyper-parameter options exist.
Popular hyper-parameter tuning frameworks like \textit{KerasTuner (KT)} and \textit{Optuna} employ more efficient alternatives to grid search~\cite{keras_tuner, optuna}.
This includes bayesian search (makes use of Gaussian priors to choose good combinations)~\cite{Mockus1989Bayesian}, \textit{random search} (randomly samples the search space)~\cite{randomSearch}, or \textit{hyperband tuning} (a variation of random search that chooses individuals after half of the iterations)~\cite{keras_tuner}.
They can sample different batches, batch sizes, learning/regularization rates, and even NN layers/units in order to find the best architecture for the task at hand.
They often find good sets of hyper-parameters within a constant amount of time as opposed to grid search's brute force method~\cite{EAGSRS}.
However, these methods do not allow for dynamic hyper-parameter optimization; each run is independent of progress made in previous runs, and most algorithms simply return hyper-parameters that the model should be initialized with during training.
Furthermore, performance on the test-set deteriorates during later training epochs, depicting the well-known overfitting curve.

One common approach for adding dynamicity to hyper-parameters is through the use of schedules~\cite{usingLRscheduleForDeepLearning}.
Learning rate schedules are often defined by three to five parameters and have been proposed to improve convergence speed~\cite{LRscheduleforfastergraddescent, GDequilibratedLR}.
These approaches are fundamentally limited as they are based on predictions about training behavior, rather than a model's actual loss.
They operate "blindly", that is, only as a function of gradient steps taken.
Such works explore cyclical, cosine-based, or random-restart schedules, adjusting the learning rate at every epoch~\cite{LRtypesadvanced}.
They introduce extra hyper-parameters that need to be searched for, causing many researchers to instead use static schedules.

\textbf{Meta-Learning} aims to find generalized models and hyper-parameters that can improve the speed of optimization on \emph{future} tasks.
They therefore attack a different problem formulation than that of hyper-parameter search, and have their own limitations \cite{PSOslow, metabad}.
Many of these techniques include hyper-parameters which require tuning themselves.
The body of work on hyper-parameter tuning, and therefore $\pd$, remains relevant even with the existence of meta-learning.

\textbf{Memetic algorithms} take advantage of both global and local optimization.
They are increasingly being used in supervised learning settings~\cite{Moscato1999MemeticAA, extraMutation, hybridResponseFiltering, Xue2021EvolutionaryGD, DANGELO2021136}.
Evolutionary stochastic gradient descent (ESGD)~\cite{ESGD} is the state-of-the-art memetic algorithm, ESGD utilizes Gaussian mutations for model parameters using an $m$-elitist average strategy to choose the best models after randomization, and SGD optimization for local search.
Performing well on CIFAR-10 classification tests, ESGD is a prime example of how adding stochastic noise benefits a strong local optimizer, because models explore more of the loss space.
Nonetheless, memetic algorithms are neither currently used nor are they designed for hyper-parameter tuning.
State-of-the-art memetic algorithms like ESGD tend to have many hyper-parameters of their own, both global (ie. mutation strength, population size, recombination probability, etc.)
This makes ESGD unfit to be used as a hyper-parameter search method, as its performance on machine learning tasks is sensitive to its own hyper-parameters.
By choosing a straightforward randomization scheme and sticking to a simple selection process $\pd$ provides two main benefits from \textbf{1)} a lower implementation complexity, and \textbf{2)} a lower number of hyper-parameters with lower sensitivity to modifications.
$\pd$ also converges faster than ESGD even after finding the best hyper-parameters for ESGD, (\textit{Figure \ref{figure:Fprog}}).
$\pd$ is thus more directly competing with hyper-parameter tuning methods rather than memetic algorithms.

\section{Conclusion}\label{conclusion}
In this paper, we propose $\pd$, a memetic algorithm that acts as a hyper-parameter tuning framework using population-based evolution.
$\pd$ actively tunes hyper-parameters during training based on a model's progress, unlike existing hyper-parameter tuners and schedules.
$\pd$'s simplicity allows it to be less sensitive to its own initial parameters.
Four extensive experiments over common supervised learning vision benchmarks demonstrate the effectiveness of $\pd$'s features in beating the combination of hyper-parameter search methods with various schedules.

\section{Reproducibility Statement}
We take many efforts to make sure that our experiments can be reevaluated effectively:
\begin{itemize}
    \item We use the number of gradient steps as our metric of "time", so that these values remain independent of the computational hardware available.
    \item We always seed every experiment taken, and those seeds are available in our source-code.
    \item We provide our reference anonymized implementation of $\pd$ and supplementary material at \url{https://github.com/anonymous112234/PopDescentAnonymous.git}.
    \item We provide a flake.nix file which exactly pins the versions of all the packages used in our tests.
\end{itemize}

\section{Impact Statement}
This paper presents work whose goal is to advance the field of Machine Learning. There are many potential societal consequences of our work, none which we feel must be specifically highlighted here.

\bibliography{citations}

\begin{thebibliography}{34}
\providecommand{\natexlab}[1]{#1}
\providecommand{\url}[1]{\texttt{#1}}
\expandafter\ifx\csname urlstyle\endcsname\relax
  \providecommand{\doi}[1]{doi: #1}\else
  \providecommand{\doi}{doi: \begingroup \urlstyle{rm}\Url}\fi

\bibitem[Akiba et~al.(2019)Akiba, Sano, Yanase, Ohta, and Koyama]{optuna}
Akiba, T., Sano, S., Yanase, T., Ohta, T., and Koyama, M.
\newblock Optuna: A next-generation hyperparameter optimization framework, 2019.

\bibitem[B\"{a}ck(1996)]{EA}
B\"{a}ck, T.
\newblock \emph{Evolutionary Algorithms in Theory and Practice: Evolution Strategies, Evolutionary Programming, Genetic Algorithms}.
\newblock Oxford University Press, Inc., USA, 1996.
\newblock ISBN 0195099710.

\bibitem[Bergstra et~al.(2011)Bergstra, Bardenet, Bengio, and K\'{e}gl]{randomSearch}
Bergstra, J., Bardenet, R., Bengio, Y., and K\'{e}gl, B.
\newblock Algorithms for hyper-parameter optimization.
\newblock 24, 2011.
\newblock URL \url{https://proceedings.neurips.cc/paper_files/paper/2011/file/86e8f7ab32cfd12577bc2619bc635690-Paper.pdf}.

\bibitem[Bergstra et~al.(2015)Bergstra, Komer, Eliasmith, Yamins, and Cox]{hyperopt}
Bergstra, J., Komer, B., Eliasmith, C., Yamins, D., and Cox, D.~D.
\newblock Hyperopt: a python library for model selection and hyperparameter optimization.
\newblock \emph{Computational Science \& Discovery}, 8\penalty0 (1):\penalty0 014008, jul 2015.
\newblock \doi{10.1088/1749-4699/8/1/014008}.
\newblock URL \url{https://dx.doi.org/10.1088/1749-4699/8/1/014008}.

\bibitem[Borna \& Hashemi(2014)Borna and Hashemi]{extraMutation}
Borna, K. and Hashemi, V.~H.
\newblock An improved genetic algorithm with a local optimization strategy and an extra mutation level for solving traveling salesman problem.
\newblock \emph{CoRR}, abs/1409.3078, 2014.
\newblock URL \url{http://arxiv.org/abs/1409.3078}.

\bibitem[Brea et~al.(2019)Brea, Simsek, Illing, and Gerstner]{saddles}
Brea, J., Simsek, B., Illing, B., and Gerstner, W.
\newblock Weight-space symmetry in deep networks gives rise to permutation saddles, connected by equal-loss valleys across the loss landscape.
\newblock \emph{CoRR}, abs/1907.02911, 2019.
\newblock URL \url{http://arxiv.org/abs/1907.02911}.

\bibitem[Choromanska et~al.(2015)Choromanska, Henaff, Mathieu, Arous, and LeCun]{choromanska2015loss}
Choromanska, A., Henaff, M., Mathieu, M., Arous, G.~B., and LeCun, Y.
\newblock The loss surfaces of multilayer networks, 2015.

\bibitem[Cui et~al.(2018)Cui, Zhang, T{\"{u}}ske, and Picheny]{ESGD}
Cui, X., Zhang, W., T{\"{u}}ske, Z., and Picheny, M.
\newblock Evolutionary stochastic gradient descent for optimization of deep neural networks.
\newblock \emph{CoRR}, abs/1810.06773, 2018.
\newblock URL \url{http://arxiv.org/abs/1810.06773}.

\bibitem[Darken et~al.(1992)Darken, Chang, and Moody]{LRscheduleforfastergraddescent}
Darken, C., Chang, J., and Moody, J.
\newblock Learning rate schedules for faster stochastic gradient search.
\newblock In \emph{Neural Networks for Signal Processing II Proceedings of the 1992 IEEE Workshop}, pp.\  3--12, Aug 1992.
\newblock \doi{10.1109/NNSP.1992.253713}.

\bibitem[Dauphin et~al.(2015)Dauphin, De~Vries, and Bengio]{GDequilibratedLR}
Dauphin, Y., De~Vries, H., and Bengio, Y.
\newblock Equilibrated adaptive learning rates for non-convex optimization.
\newblock \emph{Advances in neural information processing systems}, 28, 2015.

\bibitem[Dauphin et~al.(2014)Dauphin, Pascanu, Gulcehre, Cho, Ganguli, and Bengio]{GDsaddlePoints}
Dauphin, Y.~N., Pascanu, R., Gulcehre, C., Cho, K., Ganguli, S., and Bengio, Y.
\newblock Identifying and attacking the saddle point problem in high-dimensional non-convex optimization.
\newblock In \emph{Proceedings of the 27th International Conference on Neural Information Processing Systems - Volume 2}, NIPS'14, pp.\  2933–2941, Cambridge, MA, USA, 2014. MIT Press.

\bibitem[D’Angelo \& Palmieri(2021)D’Angelo and Palmieri]{DANGELO2021136}
D’Angelo, G. and Palmieri, F.
\newblock Gga: A modified genetic algorithm with gradient-based local search for solving constrained optimization problems.
\newblock \emph{Information Sciences}, 547:\penalty0 136--162, 2021.
\newblock ISSN 0020-0255.
\newblock \doi{https://doi.org/10.1016/j.ins.2020.08.040}.
\newblock URL \url{https://www.sciencedirect.com/science/article/pii/S0020025520308069}.

\bibitem[Gad(2022)]{PSOslow}
Gad, A.~G.
\newblock Particle {Swarm} {Optimization} {Algorithm} and {Its} {Applications}: {A} {Systematic} {Review}.
\newblock \emph{Archives of Computational Methods in Engineering}, 29\penalty0 (5):\penalty0 2531--2561, August 2022.
\newblock ISSN 1134-3060, 1886-1784.
\newblock \doi{10.1007/s11831-021-09694-4}.
\newblock URL \url{https://link.springer.com/10.1007/s11831-021-09694-4}.

\bibitem[Goffin(1977)]{inverseTime}
Goffin, J.-L.
\newblock On convergence rates of subgradient optimization methods.
\newblock \emph{Mathematical Programming}, 13:\penalty0 329--347, 1977.
\newblock URL \url{https://api.semanticscholar.org/CorpusID:37407335}.

\bibitem[Goh(2017)]{AdamMomentum}
Goh, G.
\newblock Why momentum really works.
\newblock \emph{Distill}, 2017.
\newblock \doi{10.23915/distill.00006}.
\newblock URL \url{http://distill.pub/2017/momentum}.

\bibitem[Gupta et~al.(2018)Gupta, Koren, and Singer]{Shampoo}
Gupta, V., Koren, T., and Singer, Y.
\newblock Shampoo: Preconditioned stochastic tensor optimization, 2018.

\bibitem[Kleinberg et~al.(2018)Kleinberg, Li, and Yuan]{SGD}
Kleinberg, R., Li, Y., and Yuan, Y.
\newblock An alternative view: When does sgd escape local minima?, 2018.

\bibitem[Krizhevsky(2009)]{cifar-100}
Krizhevsky, A.
\newblock Learning multiple layers of features from tiny images., 2009.

\bibitem[Krizhevsky et~al.(2009)Krizhevsky, Nair, and Hinton]{cifar}
Krizhevsky, A., Nair, V., and Hinton, G.
\newblock Cifar-10 (canadian institute for advanced research).
\newblock 2009.
\newblock URL \url{http://www.cs.toronto.edu/~kriz/cifar.html}.

\bibitem[Li et~al.(2017)Li, Tai, and E]{li2017stochastic}
Li, Q., Tai, C., and E, W.
\newblock Stochastic modified equations and adaptive stochastic gradient algorithms, 2017.

\bibitem[Li \& Arora(2019{\natexlab{a}})Li and Arora]{exponential}
Li, Z. and Arora, S.
\newblock An exponential learning rate schedule for deep learning.
\newblock \emph{CoRR}, abs/1910.07454, 2019{\natexlab{a}}.
\newblock URL \url{http://arxiv.org/abs/1910.07454}.

\bibitem[Li \& Arora(2019{\natexlab{b}})Li and Arora]{usingLRscheduleForDeepLearning}
Li, Z. and Arora, S.
\newblock An exponential learning rate schedule for deep learning.
\newblock \emph{CoRR}, abs/1910.07454, 2019{\natexlab{b}}.
\newblock URL \url{http://arxiv.org/abs/1910.07454}.

\bibitem[Liashchynskyi \& Liashchynskyi(2019)Liashchynskyi and Liashchynskyi]{EAGSRS}
Liashchynskyi, P. and Liashchynskyi, P.
\newblock Grid search, random search, genetic algorithm: {A} big comparison for {NAS}.
\newblock \emph{CoRR}, abs/1912.06059, 2019.
\newblock URL \url{http://arxiv.org/abs/1912.06059}.

\bibitem[Liu \& Theodorou(2019)Liu and Theodorou]{LRscheduleBetter}
Liu, G.-H. and Theodorou, E.~A.
\newblock Deep learning theory review: An optimal control and dynamical systems perspective, 2019.

\bibitem[Mockus(1989)]{Mockus1989Bayesian}
Mockus, J.
\newblock The bayesian approach to local optimization.
\newblock pp.\  125--156, 1989.
\newblock \doi{10.1007/978-94-009-0909-0_7}.
\newblock URL \url{https://doi.org/10.1007/978-94-009-0909-0_7}.

\bibitem[Moscato(1999)]{Moscato1999MemeticAA}
Moscato, P.
\newblock Memetic algorithms: a short introduction.
\newblock 1999.
\newblock URL \url{https://api.semanticscholar.org/CorpusID:57168143}.

\bibitem[Patacchiola et~al.(2023)Patacchiola, Sun, Hofmann, and Turner]{metabad}
Patacchiola, M., Sun, M., Hofmann, K., and Turner, R.~E.
\newblock Comparing the efficacy of fine-tuning and meta-learning for few-shot policy imitation, 2023.

\bibitem[Rogachev \& Melikhova(2020)Rogachev and Melikhova]{keras_tuner}
Rogachev, A.~F. and Melikhova, E.~V.
\newblock Automation of the process of selecting hyperparameters for artificial neural networks for processing retrospective text information.
\newblock 577\penalty0 (1):\penalty0 012012, sep 2020.
\newblock \doi{10.1088/1755-1315/577/1/012012}.
\newblock URL \url{https://dx.doi.org/10.1088/1755-1315/577/1/012012}.

\bibitem[Wolpert \& Macready(1997)Wolpert and Macready]{nofreelunch}
Wolpert, D. and Macready, W.
\newblock No free lunch theorems for optimization.
\newblock \emph{IEEE Transactions on Evolutionary Computation}, 1\penalty0 (1):\penalty0 67--82, 1997.
\newblock \doi{10.1109/4235.585893}.

\bibitem[Wu et~al.(2019)Wu, Liu, Bae, Chow, Iyengar, Pu, Wei, Yu, and Zhang]{LRtypesadvanced}
Wu, Y., Liu, L., Bae, J., Chow, K.~H., Iyengar, A., Pu, C., Wei, W., Yu, L., and Zhang, Q.
\newblock Demystifying learning rate polices for high accuracy training of deep neural networks.
\newblock \emph{CoRR}, abs/1908.06477, 2019.
\newblock URL \url{http://arxiv.org/abs/1908.06477}.

\bibitem[Xiao et~al.(2017)Xiao, Rasul, and Vollgraf]{fmnist}
Xiao, H., Rasul, K., and Vollgraf, R.
\newblock \text{Fashion-MNIST}: a novel image dataset for benchmarking machine learning algorithms, 2017.

\bibitem[Xue et~al.(2021)Xue, Qian, Xu, and Fei]{Xue2021EvolutionaryGD}
Xue, K., Qian, C., Xu, L., and Fei, X.
\newblock Evolutionary gradient descent for non-convex optimization.
\newblock In \emph{International Joint Conference on Artificial Intelligence}, 2021.
\newblock URL \url{https://api.semanticscholar.org/CorpusID:237101090}.

\bibitem[Yuenyong \& Nishihara(2014)Yuenyong and Nishihara]{hybridResponseFiltering}
Yuenyong, S. and Nishihara, A.
\newblock A hybrid gradient-based and differential evolution algorithm for infinite impulse response adaptive filtering.
\newblock \emph{International Journal of Adaptive Control and Signal Processing}, 28\penalty0 (10):\penalty0 1054--1064, 2014.
\newblock \doi{https://doi.org/10.1002/acs.2427}.
\newblock URL \url{https://onlinelibrary.wiley.com/doi/abs/10.1002/acs.2427}.

\bibitem[Zeiler(2012)]{adadelta}
Zeiler, M.~D.
\newblock Adadelta: An adaptive learning rate method, 2012.

\end{thebibliography}
\bibliographystyle{icml2024}

\clearpage
\appendix
\onecolumn
\section{Appendix}\label{appendix}

\begin{table*}[h!]
\centering
\caption{\textbf{Benchmark comparison.}
We compare different hyper-parameter tuning methods and their results on the FMNIST, CIFAR-10, and CIFAR-100 vision benchmarks, with and without regularization.
The rightmost column shows how many gradient steps each method uses to complete training.
Explanation is in \textit{Section~\ref{eval:bench}}, and the same data is plotted in \textit{Figure~\ref{fig:Benchmarks}}.}
\label{table:Bench}
\vspace*{2mm}
\begin{tabular}{c|c|c|c} 
\toprule
{\textit{Algorithm}} & {$\textit{Test Loss} \pm \sigma$} & {$\textit{Train Loss} \pm \sigma$} & {\textit{Gradient Steps}} \\ 
\midrule
\multicolumn{4}{c}{\textbf{FMNIST Without Regularization}} \\
\midrule
\textbf{Sklearn Hyperopt}     & $0.411 \pm 0.021$ &  $0.343 \pm 0.030$  & 32,000 \\
\textbf{KT RandomSearch}     & $0.277 \pm 0.023$ &  $0.112 \pm 0.034$  & 46,800 \\
\textbf{ESGD}                &  $0.276 \pm 0.009$ &  $0.114 \pm 0.007$ & 46,800 \\
\textbf{Scheduling}     & $0.270 \pm 0.036$ &  $0.149 \pm 0.068$  & 46,800 \\
\textbf{Basic Grid Search}   & $0.251 \pm 0.010$ &  $\mathbf{0.037  \pm 0.006}$ & 64,000 \\
\textbf{Population Descent}  & $\mathbf{0.249 \pm 0.020}$ &  $0.124 \pm 0.052$ & \textbf{32,000} \\
\midrule
\multicolumn{4}{c}{\textbf{FMNIST With Regularization}} \\
\midrule
\textbf{Sklearn Hyperopt}     & $0.775 \pm 0.152$ &  $0.728 \pm 0.154$  & 32,000 \\
\textbf{KT RandomSearch}     & $0.400 \pm 0.061$ &  $0.295 \pm 0.077$  & 46,800 \\
\textbf{Scheduling}     & $0.348 \pm 0.084$ &  $0.269 \pm 0.109$  & 46,800 \\
\textbf{Basic Grid Search}   &  $0.309 \pm 0.009$ &  $0.251 \pm 0.007$ & 160,000 \\
\textbf{Population Descent}  &  $\mathbf{0.262 \pm 0.019}$ & $\mathbf{0.152 \pm 0.033}$ & \textbf{32,000} \\


\midrule
\midrule
\multicolumn{4}{c}{\textbf{CIFAR-10 Without Regularization}} \\
\midrule
\textbf{Sklearn Hyperopt}     & $2.10 \pm 0.318$ &  $2.09 \pm 0.342$  & 26,000 \\
\textbf{KT RandomSearch}         & $1.51 \pm 0.275$ &  $1.343 \pm 0.296$  & 39,000 \\
\textbf{Basic Grid Search}   & $1.18 \pm 0.182$ &  $1.052  \pm 0.250$ & \textbf{19,200} \\
\textbf{ESGD}                &  $0.998 \pm 0.025$ &  $0.966 \pm 0.033$ & 93,750 \\
\textbf{Scheduling}     & $0.920 \pm 0.086$ &  $0.582 \pm 0.203$  & 39,000 \\
\textbf{Population Descent}  & $\mathbf{0.863 \pm 0.014}$ &  $\mathbf{0.577 \pm 0.060}$ & 25,600 \\
\midrule
\multicolumn{4}{c}{\textbf{CIFAR-10 With Regularization}} \\
\midrule
\textbf{Sklearn Hyperopt}     & $1.87 \pm 0.555$ &  $1.83 \pm 0.5593$  & 26,000 \\
\textbf{Scheduling}     & $1.31 \pm 0.365$ &  $1.15 \pm 0.447$  & 39,000 \\
\textbf{KT RandomSearch}         & $1.20 \pm 0.209$ &  $1.030 \pm 0.249$  & 39,000 \\
\textbf{Basic Grid Search}   &  $0.970 \pm 0.027$ &  $0.770 \pm 0.043$ & 96,000 \\
\textbf{Population Descent}  &  $\mathbf{0.843 \pm 0.030}$ &  $\mathbf{0.555 \pm 0.070}$ & \textbf{25,600} \\

\midrule
\midrule
\multicolumn{4}{c}{\textbf{CIFAR-100 Without Regularization}} \\
\midrule
\textbf{Sklearn Hyperopt}     & $4.54 \pm 0.274$ &  $4.54 \pm 0.277$  & 26,000 \\
\textbf{KT RandomSearch}         & $4.13 \pm 0.601$ &  $4.004 \pm 0.617$  & 39,000 \\
\textbf{Basic Grid Search}   & $3.43 \pm 0.050$ &  $3.30  \pm 0.041$ & 32,000 \\
\textbf{ESGD}                &  $2.88 \pm 0.146$ &  $2.74 \pm 0.157$ & 93,750 \\
\textbf{Scheduling}     & $2.68 \pm 0.363$ &  $2.31 \pm 0.523$  & 39,000 \\
\textbf{Population Descent}  & $\mathbf{2.56 \pm 0.093}$ &  $\mathbf{2.22 \pm 0.193}$ & \textbf{32,000} \\
\midrule
\multicolumn{4}{c}{\textbf{CIFAR-100 With Regularization}} \\
\midrule
\textbf{Sklearn Hyperopt}     & $4.63 \pm 0.047$ &  $4.63 \pm 0.051$  & 26,000 \\
\textbf{KT RandomSearch}         & $4.16 \pm 0.51$ &  $4.09 \pm 0.594$  & 39,000 \\
\textbf{Scheduling}     & $2.62 \pm 0.121$ &  $2.21 \pm 0.323$  & 39,000 \\
\textbf{Basic Grid Search}   &  $2.60 \pm 0.061$ &  $\mathbf{2.22 \pm 0.079}$ & 160,000 \\
\textbf{Population Descent}  &  $\mathbf{2.60 \pm 0.10}$ &  $2.27 \pm 0.176$ & \textbf{32,000} \\

\bottomrule
\end{tabular}
\end{table*}

We provide parameters to reproduce all tests conducted for our ICML 2024 submission for $\pd$, and in depth tables at the end of the appendix~\ref{table:Bench},~\ref{table:FA}.

\subsection{Benchmark Test:}

Table format of all Benchmark data is provided in \textit{Table ~\ref{table:Bench}}. All benchmark tests are run with the same parameters over 10 randomly selected seeds. We decide to train all methods until they converge, and this is how we choose how many iterations to run each of them for.

\subsubsection{FMNIST}

\textbf{CNN Architecture: } Three "Conv2D" layers with 64, 128, and 256 filters in that order. Kernel size = 3, strides = (2,2), dilation rate = (1,1), activation = "relu". Then, we use a Flatten layer.
Then, We have a fully connected layer with 1024 units (activation = "relu"), and a Dropout layer with 0.5 as its parameter, and then a fully connected (Dense) output layer with 10 units.
All parameters are the same for the without/with regularization tests, except for changing the model by adding one 12 kernel regularizer to the second to last fully connected layer in the model with regularization.
We use sparse categorical crossentropy loss.

\textbf{PopDescent:}

\textbf{Training Parameters:}

\begin{center}
\begin{tabular}{|l|l|}
\hline
\textit{Population Size} & 5 \\
\hline
\textit{Replaced Individuals (in m-elitist)} & 2 \\
\hline
\textit{Iterations} & 50 \\
\hline
\textit{Batches} & 128 \\
\hline
\textit{Batch Size} & 64 \\
\hline
\textit{lr} & 0.001 \\
\hline
\textit{Regularization Rate (in model with regularization)} & 0.001 \\
\hline
\textit{Optimizer} & Adam \\
\hline
\end{tabular}
\end{center}

\textbf{Randomization Parameters:}
\begin{center}
\begin{tabular}{|l|l|}
\hline
lr\_constant & $10^{* *}(\operatorname{random.normal}(\mathrm{mu}=-4, \operatorname{sigma}=2))$ \\
\hline
regularization\_constant & $10^{* *}(\operatorname{random.normal}(\mathrm{mu}=0, \operatorname{sigma}=2))$ \\
\hline
\begin{tabular}{l}
randomization\_amount $($ amount to randomize \\
model, changes based on model's performance) \\
\end{tabular} & $1-(2 /(2+$ model loss $))$ \\
\hline
\begin{tabular}{l}
Gaussian Noise for Model Weights (sum of current \\
weights and noise) \\
\end{tabular} & \begin{tabular}{l}
noise $=$ random.normal $(\mathrm{mu}=0$, \\
sigma=0.01)*randomization\_amount \\
\end{tabular} \\
\hline
\begin{tabular}{l}
Gaussian Noise for lr\_constant (product of \\
current lr\_constant and noise) \\
\end{tabular} & \begin{tabular}{l}
$2 * *(\mathrm{np}$. random.normal(mu=0, \\
sigma=randomization\_amount*15)) \\
\end{tabular} \\
\hline
\begin{tabular}{l}
Gaussian Noise for regularization\_rate (product \\
of current regularization\_constant and noise) \\
\end{tabular} & \begin{tabular}{l}
$2 * *(\mathrm{np}$. random.normal(mu=0, \\
sigma=randomization\_amount*15)) \\
\end{tabular} \\
\hline
\end{tabular}
\end{center}

\textbf{Basic Grid Search:} Both without/with regularization tests use the model with 12 kernel regularization, but when training without regularization, the Regularization Rates to Sample is simply [0], meaning no regularization.

\textit{Training Parameters:}

\begin{center}
\begin{tabular}{|l|l|}
\hline
Learning Rates to Sample & $[0.01,0.001,0.0001,0.00001,0.000001]$ \\
\hline
Regularization Rates to Sample & $[0.01,0.001,0.0001,0.00001,0.000001]$ \\
\hline
Iterations & 100 \\
\hline
Batches & 128 \\
\hline
Batch Size & 64 \\
\hline
Optimizer & Adam \\
\hline
\end{tabular}
\end{center}

\textbf{KT RandomSearch:} We used Keras Tuner's RandomSearch implementation.

\textit{Training Parameters:}
\begin{center}
\begin{tabular}{|l|l|}
\hline
Range of Learning Rates to Sample (continuous) & $[0.0001,0.01]$ \\
\hline
Range of Regularization Rates to Sample & $[0.00001,0.1]$ \\
\hline
max\_trials & 25 \\
\hline
executions\_per\_trial & 2 \\
\hline
\begin{tabular}{l}
train\_epochs (with early stopping, patience=2; \\
train over the whole train dataset) \\
\end{tabular} & 20 \\
\hline
Batch Size & 64 \\
\hline
Optimizer & Adam \\
\hline
\end{tabular}
\end{center}

\textbf{KT Scheduling:} We used Keras Tuner’s RandomSearch implementation to choose from three different Keras learning rate schedules (exponential decay, inverse time decay, polynomial decay), and searched for their parameters as well.

\textit{Exponential Decay Parameters:}

\begin{center}
\begin{tabular}{|l|l|}
\hline
\begin{tabular}{l}
Range of Learning Rates to Sample (continuous, \\
$\log )$ \\
\end{tabular} & $[0.0001,0.01]$ \\
\hline
Range of Regularization Rates to Sample & $[0.00001,0.1]$ \\
\hline
Range of Decay Rates to Sample & $[0.8,0.99]$ \\
\hline
Range of Decay Steps to Sample & $[1000,10000]$ \\
\hline
\end{tabular}
\end{center}

\textit{Inverse Time Decay Parameters:}

\begin{center}
\begin{tabular}{|l|l|}
\hline
\begin{tabular}{l}
Range of Learning Rates to Sample (continuous, \\
$\log )$ \\
\end{tabular} & $[0.0001,0.01]$ \\
\hline
Range of Regularization Rates to Sample & $[0.00001,0.1]$ \\
\hline
Range of Decay Rates to Sample & $[0.8,0.99]$ \\
\hline
Range of Decay Steps to Sample & $[1000,10000]$ \\
\hline
\end{tabular}
\end{center}

\textit{Polynomial Decay Parameters:}

\begin{center}
\begin{tabular}{|l|l|}
\hline
\begin{tabular}{l}
Range of Initial Learning Rates to Sample \\
(continuous, log) \\
\end{tabular} & $[0.0001,0.01]$ \\
\hline
\begin{tabular}{l}
Range of End Learning Rates to Sample \\
(continuous, log) \\
\end{tabular} & $[0.00001,0.01]$ \\
\hline
Range of Regularization Rates to Sample & $[0.00001,0.1]$ \\
\hline
Range of Decay Rates to Sample & $[0.8,0.99]$ \\
\hline
Range of Decay Steps to Sample & $[1000,10000]$ \\
\hline
Range of "Power" Term to Sample & $[0.1,2]$ \\
\hline
\end{tabular}
\end{center}

\textit{Search Parameters}

\begin{center}
\begin{tabular}{|l|l|}
\hline
max\_trials & 25 \\
\hline
executions\_per\_trial & 2 \\
\hline
\begin{tabular}{l}
train\_epochs (with early stopping, patience=0; \\
train over the whole train dataset) \\
\end{tabular} & 20 \\
\hline
Batch Size & 64 \\
\hline
Optimizer & Adam \\
\hline
\end{tabular}
\end{center}

\textbf{Sklearn Hyperopt:} We used Sklearn Hyperopt \textbf{TPE}, their version of Bayesian Search.

\textit{Training + Search Parameters}

\begin{center}
\begin{tabular}{|l|l|}
\hline
max\_trials & 32 \\
\hline
executions\_per\_trial & 1 \\
\hline
\begin{tabular}{l}
train\_epochs (with early stopping, patience=0; \\
train over the whole train dataset) \\
\end{tabular} & 20 \\
\hline
Batch Size & 64 \\
\hline
Optimizer & Adam \\
\hline
\end{tabular}
\end{center}

\textbf{ESGD:} We used $t q c h$ 's open-source implementation of ESGD on \href{https://github.com/tqch/esgd-ws}{https://github.com/tqch/esgd-ws}.

\textit{Training Parameters:}

\begin{center}
\begin{tabular}{|l|l|}
\hline
$n \_p o p u l a t i o n$ & 5 \\
\hline
sgds\_per\_gen & 1 \\
\hline
evos\_per\_gen & 1 \\
\hline
reproductive\_factor & 4 \\
\hline
m\_elite & 3 \\
\hline
mixing\_number & 3 \\
\hline
optimizer\_class & SGD $(\mathrm{lr}=0.001)$ \\
\hline
n\_generations & 10 \\
\hline
batch\_size & 64 \\
\hline
\end{tabular}
\end{center}

\subsubsection{CIFAR-10/CIFAR-100}

\textbf{CNN Architecture: } Two "Conv2D" layers with 32, and 64 filters in that order. Then, a "MaxPooling2D" layer with (2,2) as its parameter, a third Conv2D layer with 128 filters, and another MaxPooling2D layer with (2,2) as its parameter.
We add a final Conv2D layer with 64 filters, and a last MaxPooling2D layer with (4,4) as its parameter.
All Conv2D layers use Kernel size = 3, strides = (2,2), dilation rate = (1,1), activation = "relu".
Then, we have a Flatten layer, followed by a fully connected (Dense) layer with 256 units, and a final fully connected layer for the output.
All parameters are the same for the without/with regularization tests, except for changing the model by adding one 12 kernel regularizer to the second to last fully connected layer in the model with regularization.
We use sparse categorical crossentropy loss.
\textit{All parameters/models for CIFAR-10 and CIFAR-100 are the exact same except for the last layer in CIFAR-100 being adjusted to output 100 classes instead of 10.}

\textbf{PopDescent:}

\textit{Training Parameters:}

\begin{center}
\begin{tabular}{|l|l|}
\hline
Population Size & 5 \\
\hline
Replaced Individuals (in m-elitist) & 2 \\
\hline
Iterations & 20 \\
\hline
Batches (trained over 2 epochs per iteration here) & 128 \\
\hline
Batch Size & 64 \\
\hline
lr & 0.001 \\
\hline
Regularization Rate (in model with regularization) & 0.001 \\
\hline
Optimizer & Adam \\
\hline
\end{tabular}
\end{center}

\textit{Randomization Parameters:}

\begin{center}
\begin{tabular}{|c|c|}
\hline
lr\_constant & $10 * *(\operatorname{random} \cdot \operatorname{normal}(\mathrm{mu}=-4, \operatorname{sigma}=2))$ \\
\hline
regularization\_constant & $10 * *(\operatorname{random} \cdot \operatorname{normal}(\mathrm{mu}=0, \operatorname{sigma}=2))$ \\
\hline
\begin{tabular}{l}
randomization\_amount (amount to randomize \\
model, changes based on model's performance) \\
\end{tabular} & $1-(2 /(2+$ model loss $))$ \\
\hline
\begin{tabular}{l}
Gaussian Noise for Model Weights (sum of current \\
weights and noise) \\
\end{tabular} & \begin{tabular}{l}
noise $=$ random.normal $(\mathrm{mu}=0$, \\
sigma $=0.01) *$ randomization\_amount \\
\end{tabular} \\
\hline
\begin{tabular}{l}
Gaussian Noise for lr\_constant (product of \\
current lr\_constant and noise) \\
\end{tabular} & \begin{tabular}{l}
$2 * *(\mathrm{np}$. random.normal $(\mathrm{mu}=0$ \\
sigma=randomization\_amount*15)) \\
\end{tabular} \\
\hline
\begin{tabular}{l}
Gaussian Noise for regularization\_rate (product \\
of current regularization\_constant and noise) \\
\end{tabular} & \begin{tabular}{l}
$2 * *($ np.random.normal $(\mathrm{mu}=0$ \\
sigma=randomization\_amount*15)) \\
\end{tabular} \\
\hline
\end{tabular}
\end{center}

\textbf{Basic Grid Search:} Both without/with regularization tests use the model with 12 kernel regularization, but when training without regularization, the Regularization Rates to Sample is simply [0], meaning no regularization.

\textit{Training Parameters:}

\begin{center}
\begin{tabular}{|l|l|}
\hline
Learning Rates to Sample & $[0.01,0.001,0.0001,0.00001,0.000001]$ \\
\hline
Regularization Rates to Sample & $[0.01,0.001,0.0001,0.00001,0.000001]$ \\
\hline
Iterations & 30 \\
\hline
Batches & 128 \\
\hline
Batch Size & 64 \\
\hline
Optimizer & Adam \\
\hline
\end{tabular}
\end{center}

\textbf{KT RandomSearch:} We used Keras Tuner's RandomSearch implementation.

\textit{Training Parameters:}

\begin{center}
\begin{tabular}{|l|l|}
\hline
Range of Learning Rates to Sample (continuous) & $[0.0001,0.01]$ \\
\hline
Range of Regularization Rates to Sample & $[0.00001,0.1]$ \\
\hline
max\_trials & 25 \\
\hline
executions\_per\_trial & 2 \\
\hline
\begin{tabular}{l}
train\_epochs (with early stopping, patience=2; \\
train over the whole train dataset) \\
\end{tabular} & 20 \\
\hline
Batch Size & 64 \\
\hline
Optimizer & Adam \\
\hline
\end{tabular}
\end{center}

\textbf{KT Scheduling:}  We used Keras Tuner’s RandomSearch implementation to choose from three different Keras learning rate schedules (exponential decay, inverse time decay, polynomial decay), and search for their parameters as well.

\textit{Exponential Decay Parameters:}

\begin{center}
\begin{tabular}{|l|l|}
\hline
\begin{tabular}{l}
Range of Learning Rates to Sample (continuous, \\
$\log )$ \\
\end{tabular} & $[0.0001,0.01]$ \\
\hline
Range of Regularization Rates to Sample & $[0.00001,0.1]$ \\
\hline
Range of Decay Rates to Sample & $[0.8,0.99]$ \\
\hline
Range of Decay Steps to Sample & $[1000,10000]$ \\
\hline
\end{tabular}
\end{center}

\textit{Inverse Time Decay Parameters:}

\begin{center}
\begin{tabular}{|l|l|}
\hline
\begin{tabular}{l}
Range of Learning Rates to Sample (continuous, \\
$\log )$ \\
\end{tabular} & $[0.0001,0.01]$ \\
\hline
Range of Regularization Rates to Sample & $[0.00001,0.1]$ \\
\hline
Range of Decay Rates to Sample & $[0.8,0.99]$ \\
\hline
Range of Decay Steps to Sample & $[1000,10000]$ \\
\hline
\end{tabular}
\end{center}

\textit{Polynomial Decay Parameters:}

\begin{center}
\begin{tabular}{|l|l|}
\hline
\begin{tabular}{l}
Range of Initial Learning Rates to Sample \\
(continuous, log) \\
\end{tabular} & $[0.0001,0.01]$ \\
\hline
\begin{tabular}{l}
Range of End Learning Rates to Sample \\
(continuous, log) \\
\end{tabular} & $[0.00001,0.01]$ \\
\hline
Range of Regularization Rates to Sample & $[0.00001,0.1]$ \\
\hline
Range of Decay Rates to Sample & $[0.8,0.99]$ \\
\hline
Range of Decay Steps to Sample & $[1000,10000]$ \\
\hline
Range of "Power" Term to Sample & $[0.1,2]$ \\
\hline
\end{tabular}
\end{center}

\textit{Search Parameters}

\begin{center}
\begin{tabular}{|l|l|}
\hline
max\_trials & 25 \\
\hline
executions\_per\_trial & 2 \\
\hline
\begin{tabular}{l}
train\_epochs (with early stopping, patience=0; \\
train over the whole train dataset) \\
\end{tabular} & 20 \\
\hline
Batch Size & 64 \\
\hline
Optimizer & Adam \\
\hline
\end{tabular}
\end{center}

\textbf{Sklearn Hyperopt:} We used Sklearn Hyperopt \textbf{TPE}, their version of Bayesian Search.

\textit{Training + Search Parameters}

\begin{center}
\begin{tabular}{|l|l|}
\hline
max\_trials & 37 \\
\hline
executions\_per\_trial & 1 \\
\hline
\begin{tabular}{l}
train\_epochs (with early stopping, patience=0; \\
train over the whole train dataset) \\
\end{tabular} & 20 \\
\hline
Batch Size & 64 \\
\hline
Optimizer & Adam \\
\hline
\end{tabular}
\end{center}

\textbf{ESGD:} We used $t q c h$ 's open-source implementation of ESGD on \href{https://github.com/tqch/esgd-ws}{https://github.com/tqch/esgd-ws}.

\textit{Training Parameters:}

\begin{center}
\begin{tabular}{|l|l|}
\hline
$n \_p o p u l a t i o n$ & 5 \\
\hline
sgds\_per\_gen & 1 \\
\hline
evos\_per\_gen & 1 \\
\hline
reproductive\_factor & 4 \\
\hline
m\_elite & 3 \\
\hline
mixing\_number & 3 \\
\hline
optimizer\_class & SGD (lr=0.001) \\
\hline
n\_generations & 3 \\
\hline
batch\_size & 8 \\
\hline
\end{tabular}
\end{center}

\subsection{Convergence Test}

All convergence tests are run with the same parameters over 6 randomly selected seeds. All tests are run on the same model without regularization on the FMNIST dataset.

\begin{table*}[h]
\centering
\caption{\textbf{Convergence Test.}
We run each algorithm on six random seeds, and track their validation loss convergence on the FMNIST dataset
All methods are tuning the learning rate, without regularization.
A graph format of the same convergence data is seen in \textit{Figure~\ref{figure:Fprog}}.
Here, we provide the final validation loss each method achieves after 50k gradient steps.}
\label{table:Convergence}
\vspace*{2mm}
\begin{tabular}{c|c} 
\toprule
{\textbf{Algorithm}} & {\textbf{Final Val-Loss EMA}} \\ 
\midrule
\textit{$\pd$} & 0.260 \\
\midrule
\textit{Grid Search} & 0.611 \\
\midrule
\textit{KT RandomSearch} & 0.451 \\
\midrule
\textit{KT Scheduling} & 0.292 \\
\midrule
\textit{Sklearn Hyperopt TPE} & 0.433 \\
\midrule
\textit{ESGD} &  0.336 \\
\midrule
\end{tabular}
\end{table*}



\textbf{PopDescent:}

\textit{Training Parameters:}
\begin{center}
\begin{tabular}{|l|l|}
\hline
Population Size & 5 \\
\hline
Replaced Individuals (in m-elitist) & 2 \\
\hline
Iterations & 115 \\
\hline
Batches & 128 \\
\hline
Batch Size & 64 \\
\hline
lr & 0.001 \\
\hline
Optimizer & Adam \\
\hline
\end{tabular}
\end{center}

\textit{Randomized Parameters:}
\begin{center}
\begin{tabular}{|l|l|}
\hline
lr\_constant & $10^{* *}(\operatorname{random.normal}(\mathrm{mu}=-4, \operatorname{sigma}=2))$ \\
\hline
\begin{tabular}{l}
randomization\_amount (amount to randomize \\
model, changes based on model's performance) \\
\end{tabular} & $1-(2 /(2+\operatorname{model} \operatorname{loss}))$ \\
\hline
\begin{tabular}{l}
Gaussian Noise for Model Weights (sum of current \\
weights and noise) \\
\end{tabular} & \begin{tabular}{l}
noise $=$ random.normal(mu=0, \\
sigma $=0.01) *$ randomization\_amount \\
\end{tabular} \\
\hline
\begin{tabular}{l}
Gaussian Noise for lr\_constant (product of \\
current lr\_constant and noise) \\
\end{tabular} & \begin{tabular}{l}
$2 * *(n p . r a n d o m . n o r m a l(m u=0$, \\
sigma=randomization\_amount*15)) \\
\end{tabular} \\
\hline
\end{tabular}
\end{center}

\textbf{Basic Grid Search:}

\textit{Training Parameters}

\begin{center}
\begin{tabular}{|l|l|}
\hline
Learning Rates to Sample & $[0.01,0.001,0.0001,0.00001,0.000001]$ \\
\hline
Regularization Rates to Sample & $[0]$ \\
\hline
Iterations & 100 \\
\hline
Batches & 128 \\
\hline
Batch Size & 64 \\
\hline
Optimizer & Adam \\
\hline
\end{tabular}
\end{center}

\textbf{KT RandomSearch:}

\textit{Training Parameters}
\begin{center}
\begin{tabular}{|l|l|}
\hline
Range of Learning Rates to Sample (continuous) & $[0.0001,0.01]$ \\
\hline
Range of Regularization Rates to Sample & $[0.00001,0.1]$ \\
\hline
max\_trials & 25 \\
\hline
executions\_per\_trial & 2 \\
\hline
\begin{tabular}{l}
train\_epochs (with early stopping, patience=2; \\
train over the whole train dataset) \\
\end{tabular} & 20 \\
\hline
Batch Size & 64 \\
\hline
Optimizer & Adam \\
\hline
\end{tabular}
\end{center}

\textbf{KT Scheduling:}

\textit{Exponential Decay Parameters:}

\begin{center}
\begin{tabular}{|l|l|}
\hline
\begin{tabular}{l}
Range of Learning Rates to Sample (continuous, \\
$\log )$ \\
\end{tabular} & $[0.0001,0.01]$ \\
\hline
Range of Regularization Rates to Sample & $[0.00001,0.1]$ \\
\hline
Range of Decay Rates to Sample & $[0.8,0.99]$ \\
\hline
Range of Decay Steps to Sample & $[1000,10000]$ \\
\hline
\end{tabular}
\end{center}

\textit{Inverse Time Decay Parameters:}

\begin{center}
\begin{tabular}{|l|l|}
\hline
\begin{tabular}{l}
Range of Learning Rates to Sample (continuous, \\
$\log )$ \\
\end{tabular} & $[0.0001,0.01]$ \\
\hline
Range of Regularization Rates to Sample & $[0.00001,0.1]$ \\
\hline
Range of Decay Rates to Sample & $[0.8,0.99]$ \\
\hline
Range of Decay Steps to Sample & $[1000,10000]$ \\
\hline
\end{tabular}
\end{center}

\textit{Polynomial Decay Parameters:}

\begin{center}
\begin{tabular}{|l|l|}
\hline
\begin{tabular}{l}
Range of Initial Learning Rates to Sample \\
(continuous, log) \\
\end{tabular} & $[0.0001,0.01]$ \\
\hline
\begin{tabular}{l}
Range of End Learning Rates to Sample \\
(continuous, log) \\
\end{tabular} & $[0.00001,0.01]$ \\
\hline
Range of Regularization Rates to Sample & $[0.00001,0.1]$ \\
\hline
Range of Decay Rates to Sample & $[0.8,0.99]$ \\
\hline
Range of Decay Steps to Sample & $[1000,10000]$ \\
\hline
Range of "Power" Term to Sample & $[0.1,2]$ \\
\hline
\end{tabular}
\end{center}

\textit{Search Parameters}

\begin{center}
\begin{tabular}{|l|l|}
\hline
max\_trials & 25 \\
\hline
executions\_per\_trial & 2 \\
\hline
\begin{tabular}{l}
train\_epochs (with early stopping, patience=0; \\
train over the whole train dataset) \\
\end{tabular} & 20 \\
\hline
Batch Size & 64 \\
\hline
Optimizer & Adam \\
\hline
\end{tabular}
\end{center}

\textbf{Sklearn Hyperopt:} All parameters are the same for the without/with regularization tests, except for changing the model by adding one 12 kernel regularizer to the second to last fully connected layer in the model with regularization. We used Sklearn Hyperopt \textbf{TPE}, their version of Bayesian Search.

\textit{Training + Search Parameters}

\begin{center}
\begin{tabular}{|l|l|}
\hline
max\_trials & 32 \\
\hline
executions\_per\_trial & 1 \\
\hline
\begin{tabular}{l}
train\_epochs (with early stopping, patience=0; \\
train over the whole train dataset) \\
\end{tabular} & 20 \\
\hline
Batch Size & 64 \\
\hline
Optimizer & Adam \\
\hline
\end{tabular}
\end{center}

\textbf{ESGD:}

\textit{Training Parameters}

\begin{center}
\begin{tabular}{|l|l|}
\hline
$n \_p o p u l a t i o n$ & 5 \\
\hline
sgds\_per\_gen & 1 \\
\hline
evos\_per\_gen & 1 \\
\hline
reproductive\_factor & 4 \\
\hline
m\_elite & 3 \\
\hline
mixing\_number & 3 \\
\hline
optimizer\_class & SGD $(\mathrm{lr}=0.001)$ \\
\hline
n\_generations & 15 \\
\hline
batch\_size & 64 \\
\hline
\end{tabular}
\end{center}

\subsection{\textbf{Ablation Study:}}

To emphasize the differences of PopDescent's features, we train only on the first 10k images in the FMNIST dataset, and add 12 kernel regularization to every layer in the benchmark FMNIST model, initialized with a default value of 0.001 in each layer. All training parameters are the same in each of the four tests listed in the paper. We only change whether PopDescent's randomization scheme is on or off, with $\mathrm{CV}$ selection turned on, and using the model with regularization for the first two tests listed ("Ablation Study Over Randomization"). We only change the selection process when comparing without $\mathrm{CV}$ selection vs with $\mathrm{CV}$ selection, both on a model without regularization for the second two tests listed ("Ablation Study Over Cross-Validation Fitness").

\textit{Training Parameters:}

\begin{center}
\begin{tabular}{|l|l|}
\hline
Population Size & 10 \\
\hline
Replaced Individuals (in m-elitist) & 5 \\
\hline
Iterations & 35 \\
\hline
Batches & 128 \\
\hline
Batch Size & 64 \\
\hline
$l r$ & 0.001 \\
\hline
Regularization Rate & 0.001 \\
\hline
Optimizer & Adam \\
\hline
\end{tabular}
\end{center}

\textit{Randomization Parameters:}
\begin{center}
\begin{tabular}{|l|l|}
\hline
Ir\_constant & $10^{* *}(\operatorname{random.normal}(\mathrm{mu}=-4, \operatorname{sigma}=2))$ \\
\hline
regularization\_constant & $10^{* *}(\operatorname{random.normal}(\mathrm{mu}=0, \operatorname{sigma}=2))$ \\
\hline
\begin{tabular}{l}
randomization\_amount $($ amount to randomize \\
model, changes based on model's performance) \\
\end{tabular} & $1-(2 /(2+$ model loss $))$ \\
\hline
\begin{tabular}{l}
Gaussian Noise for Model Weights (sum of current \\
weights and noise) \\
\end{tabular} & \begin{tabular}{l}
noise $=$ random.normal $(\mathrm{mu}=0$, \\
sigma=0.01)*randomization\_amount \\
\end{tabular} \\
\hline
\begin{tabular}{l}
Gaussian Noise for lr\_constant (product of \\
current lr\_constant and noise) \\
\end{tabular} & \begin{tabular}{l}
$2 * *(\mathrm{np}$. random.normal(mu=0, \\
sigma=randomization\_amount*15)) \\
\end{tabular} \\
\hline
\begin{tabular}{l}
Gaussian Noise for regularization\_rate (product \\
of current regularization\_constant and noise) \\
\end{tabular} & \begin{tabular}{l}
$2 * *(\mathrm{np}$. random.normal(mu=0, \\
sigma=randomization\_amount*15)) \\
\end{tabular} \\
\hline
\end{tabular}
\end{center}

\begin{table*}[h!]
\centering
\caption{\textbf{Ablation study FMNIST.}
We demonstrate how models react to features of $\pd$ being turned on v.s. turned off perform on a condensed version (10k training images) of the FMNIST dataset. Explanation can be found in Section~\ref{eval:ablation}.}
\label{table:FA}
\vspace*{2mm}
\begin{tabular}{c|c|c|c|c} 
\toprule
{\textit{Randomization}} & \textit{CV Selection} & \textit{Regularization} & {$\textit{Test Loss} \pm \sigma$} & {$\textit{Train Loss} \pm \sigma$} \\ 
\midrule
\multicolumn{5}{c}{\textbf{Ablation Study Over \pd\ Randomization}} \\
\midrule
\cmark         & \cmark & \cmark &  $0.345  \pm 0.006$ & $0.139  \pm 0.028$ \\
\xmark         & \cmark & \cmark &  $0.412 \pm 0.005$  & $0.118  \pm 0.077$ \\
\midrule
\multicolumn{5}{c}{\textbf{Ablation Study Over Cross-Validation Fitness}} \\
\midrule
\cmark         & \cmark & \xmark &  $0.356 \pm 0.009$ &  $0.163 \pm 0.019$ \\
\cmark         & \xmark  & \xmark &  $1.140 \pm 0.147$  & $0.0003 \pm 0.0002$ \\
\bottomrule
\end{tabular}
\end{table*}

\subsection{\textbf{Hyper-parameter Sensitivity Study:}}

We conduct this study on the CIFAR-10 dataset, on the same model used in the CIFAR-10 benchmark tests without regularization. When testing for the effects of changing the number of iterations vs different learning rates, we keep everything constant, except for changing the hyper-parameter we are testing by sampling each possibility in the arrays listed for that category.

\textit{Training Parameters:}
\begin{center}
\begin{tabular}{|c|c|}
\hline
Population Size & 10 \\
\hline
Replaced Individuals (in m-elitist) & 5 \\
\hline
Iterations & $[10, \mathbf{3 0}, \mathbf{5 0}]$ \\
\hline
Batches & 128 \\
\hline
Batch Size & 64 \\
\hline
lr & $[0.001,0.01,0.05]$ \\
\hline
Regularization Rate & 0.001 \\
\hline
Optimizer & Adam \\
\hline
\end{tabular}
\end{center}

\textit{Randomization Parameters:}

\begin{center}
\begin{tabular}{|l|l|}
\hline
lr\_constant & $10^{* *}(\operatorname{random.normal}(\mathrm{mu}=-4, \operatorname{sigma}=2))$ \\
\hline
regularization\_constant & $10^{* *}(\operatorname{random.normal}(\mathrm{mu}=0, \operatorname{sigma}=2))$ \\
\hline
\begin{tabular}{l}
randomization\_amount $($ amount to randomize \\
model, changes based on model's performance) \\
\end{tabular} & $1-(2 /(2+\operatorname{model} \operatorname{loss}))$ \\
\hline
\begin{tabular}{l}
Gaussian Noise for Model Weights (sum of current \\
weights and noise) \\
\end{tabular} & \begin{tabular}{l}
noise $=$ random.normal $(\mathrm{mu}=0$, \\
sigma $=0.01) *$ randomization\_amount \\
\end{tabular} \\
\hline
\begin{tabular}{l}
Gaussian Noise for lr\_constant (product of \\
current lr\_constant and noise) \\
\end{tabular} & \begin{tabular}{l}
$2^{* *}(\mathrm{np}$. random.normal(mu=0, \\
sigma=randomization\_amount*15)) \\
\end{tabular} \\
\hline
\begin{tabular}{l}
Gaussian Noise for regularization\_rate (product \\
of current regularization\_constant and noise) \\
\end{tabular} & \begin{tabular}{l}
$2 * *(\mathrm{np}$. random.normal(mu=0, \\
sigma=randomization\_amount*15)) \\
\end{tabular} \\
\hline
\end{tabular}
\end{center}

\textbf{ESGD:}

\textit{Training Parameters:}

\begin{center}
\begin{tabular}{|c|c|}
\hline
n\_population & 5 \\
\hline
sgds\_per\_gen & 1 \\
\hline
evos\_per\_gen & 1 \\
\hline
reproductive factor & 4 \\
\hline
m\_elite & 3 \\
\hline
mixing\_number & 3 \\
\hline
optimizer\_class & SGD \\
\hline
learning\_rate & $[0.001,0.01,0.05]$ \\
\hline
n\_generations & $[1,3,5]$ \\
\hline
batch\_size & 8 \\
\hline
\end{tabular}
\end{center}

\end{document}